\documentclass{ifacconf}

\usepackage{graphicx}      
\usepackage{natbib}        
\usepackage{subcaption}
\usepackage{url}
\usepackage{graphicx} %
\usepackage{multirow}
\usepackage{booktabs}
\usepackage{amsmath}
\usepackage{amssymb}  
\usepackage{subcaption}

\newcommand{\sys}{S}

\newcommand{\free}{\mathcal{M}}  
\newcommand{\D}{\mathcal{D}}

\newcommand{\nsamp}{N}
\newcommand{\norm}[1]{\left\lVert#1\right\rVert}

\newcommand{\E}{\mathbb{E}}

\begin{document}
\begin{frontmatter}

\title{On the adaptation of in-context learners for system identification}


 \author{Dario Piga,}
\author{Filippo Pura,} 
\author{Marco Forgione}

\address{IDSIA Dalle Molle Institute for Artificial Intelligence USI-SUPSI, Via la Santa 1, CH-6962 Lugano-Viganello, Switzerland.}

\begin{abstract}                
In-context  system identification  aims at constructing meta-models to describe classes of systems, differently from traditional approaches that model  single systems. This paradigm facilitates the leveraging of knowledge acquired from observing the behaviour of different, yet related dynamics. This paper discusses  the role of meta-model adaptation. Through numerical examples, we demonstrate how meta-model adaptation can enhance predictive performance in three realistic scenarios: tailoring the meta-model to describe a specific system rather than a class; extending the meta-model to capture the behaviour of systems beyond the initial training class; and recalibrating the model for new prediction tasks. Results highlight the effectiveness of meta-model adaptation to achieve a more robust and versatile meta-learning framework for system identification.
\end{abstract}


\end{frontmatter}

\section{Introduction}

In traditional system identification, the focus is on learning a model of a specific system through a-priori physical knowledge and measured input-output trajectories. 
This typical workflow is closely related to supervised machine learning, albeit with distinct peculiarities. The emphasis is on dynamical systems, opting for parsimonious representations like Linear Parameter-Varying and hybrid models \citep{bamieh2002identification, mejari2020recursive}, and employing the model for  applications such as control design \citep{bombois2006least, piga2019performance}. Given the intrinsic link between supervised learning and system identification, recent contributions employ  deep learning tools to estimate  dynamical systems using neural network structures \citep{masti2021learning,forgione2021continuous,forgione2021dynonet,beintema2023deep,pillonetto2023deep}. Despite these advancements, the conventional approach often overlooks potential insights from previous trainings on similar systems, leaving room for enhanced methodologies that leverage knowledge accumulated across related systems and tasks.

Harnessing this knowledge represents a pioneering shift for system identification and adaptive control (see,  e.g.,~\citet{zhan2022calibrating,richards2021adaptive,chakrabarty2023meta,balim2023can}), situated within the  domain of meta-learning or ``learning to learn'' \citep{schmidhuber1987evolutionary,finn2017model}. Meta-learning involves training models across various tasks to enable these models to adapt to new ones with minimal human intervention, training effort, or data. This facilitates generalization  from one task to another, a significant leap towards autonomous system identification and control.

In the in-context system identification framework introduced by the authors in \citet{forgione2023context}, instead of estimating a model of a specific dynamical system, a meta-model describing a whole class of such systems is 
learned.
Meta-model pre-training is performed using data drawn from a potentially infinite stream of input/output sequences generated from different, yet related systems.  In practice, data could be sourced from simulators, allowing for the generation of an arbitrary amount of synthetic input/output sequences by varying physical parameters and disturbances. 
  To guarantee substantial representational power of the meta-model, an encoder-decoder  Transformer architecture is adopted \citep{vaswani2017attention,radford2019language} and  specialized to process real-valued sequences. 
The Transformer acts as an in-context learner~\citep{kirsch2022general}, taking a short input-output sequence (context) and producing predictions without the need for a specialized model of the system under study. This eliminates the necessity to specify a dynamical model structure, implement a training algorithm,  tune hyper-parameters, etc.

Although \citet{forgione2023context} shows that the meta-model  exhibits high performance on test systems belonging to the same class used for training, model adaptation may be necessary to further enhance its predictive accuracy.  
Indeed, this is a need not only in meta-learning, but also in traditional system identification to take into account, e.g., aging, changing configurations, and different external conditions (see, e.g., \citet{pozzoli2020tustin, forgione2023adaptation}).

In this paper, we extend the work in~\citet{forgione2023context} showing through numerical examples that only minimal data sequences and a few iterations of gradient descent for weight updates are needed to improve the model's predictive performance. We consider three  scenarios where adaptation can be applied: refining the meta-model to describe a specific system rather than a class; adapting the meta-model to describe the behaviour of a system outside the class used for training; and modifying the model for new prediction tasks, specifically transitioning from short-term (easy) to long-term prediction (more difficult) task.

The rest of the paper is organized as follows. Section~\ref{sec:prob_Set} discusses the problem setting, elaborates on the meta-learning framework introduced in~\citet{forgione2023context}, and provides a more detailed discussion on the model adaptation scenarios addressed in this paper. Architectures and algorithms for training and adaptation are described in Section~\ref{sec:meta_sys}. Numerical examples illustrating the three  adaptation scenarios are presented in Section~\ref{sec:examples} and conclusions are drawn in Section~\ref{sec:conc}.

\section{Problem setting} \label{sec:prob_Set}

\subsection{Conventional   \emph{vs.} meta identification}

Conventional system identification seeks to estimate the model of a \emph{fixed} unknown dynamical system $\sys$ using an input-output sequence $\D = (u_{1:\nsamp}, y_{1:\nsamp})$ generated by $\sys$, where $u_k \in \mathbb{R}^{n_u}$ (resp. $y_k  \in \mathbb{R}^{n_y}$) represents the system's input (resp. output) at time step $k$.

In the in-context learning framework introduced in \citet{forgione2023context} and discussed in this paper, a \emph{probability  distribution} of dynamical systems is considered. This distribution is used to generate a sequence of such systems along with their corresponding input-output datasets. Thus, we have access to an \emph{infinite stream} of input/output pairs 
$ \{ \mathcal{D}^{(i)} = (u_{1:N}^{(i)}, y_{1:N}^{(i)}), \, i=1,2,\dots, \infty \} $, 
each derived by feeding a randomly generated signal $ u_{1:N}^{(i)}$ as input to a randomly instantiated dynamical system $\sys^{(i)}$. Formally, we can sample synthetic datasets $ \mathcal{D}^{(i)} $ from an underlying distribution $p(\mathcal{D})$, which may not be explicitly known. Since the datasets $\D^{(i)}$ are drawn from the common distribution $p(\D)$, partial knowledge transfer from one dataset to another is possible.

\subsection{Learning system classes}

As discussed in \citet{forgione2023context}, an in-context learner  $\free_{\phi}$ (called meta-model), with parameters $\phi$, can be utilized to describe a whole class of systems. The meta-model $\free_{\phi}$   processes  the input-output dataset $\D^{(i)}$ and directly provides the predictions of interest, without generating a  specialized  model of the system $\sys^{(i)}$ generating the dataset $\D^{(i)}$.

Specifically,  the meta-model $\free_{\phi}$ receives the input/output sequence ($u_{1:m}^{(i)}, y_{1:m}^{(i)}$) up to time step $m$
and a test input sequence $u_{m+1:N}^{(i)}$ from time step $m+1$ to $N$. Its goal is to produce the corresponding simulated output sequence 
$\hat y_{m+1:N}^{(i)}$: 
\begin{equation}
\label{eq:model_free_sim}
\hat y_{m+1:N}^{(i)} = \free_\phi(u_{1:m}^{(i)}, y_{1:m}^{(i)}, u_{m+1:N}^{(i)}).
\end{equation}

 To solve such a meta-learning problem,  the meta-model $ \free_\phi$ is expected to ``understand'' (to a certain degree) the data generating mechanism $\sys^{(i)}$ from the provided \emph{context}   ($u_{1:m}^{(i)}, y_{1:m}^{(i)}$), without returning an explicit representation in a model form. 

\subsection{Adaptation}
Once the meta-model $\free_{\phi}$ is estimated based on the data drawn from the probability distribution $p(\D)$, the model may need to be adapted to enhance its predictive capabilities in various scenarios, such as: 

\begin{itemize}
    \item \textbf{From class to  system:} Adapt the model to describe data not generated by the  distribution $p(\D)$, but by the conditional distribution $p(\D|\sys)$. This case encompasses the scenario where we want to specialize the model $\free_{\phi}$ not to describe a generic class of systems, but to move towards a narrower model describing a specific system $\sys$ generating data $\D$.
    
    \item \textbf{From class to class:} Adapt the model in case of a shift in the data-generating probability distribution $p(\D)$. This encompasses the scenario where the model class we aim to describe is not included in the class used for training. The class-to-system and the class-to-class problems can also be considered together  to handle a class to out-of-class-system problem. In this case, a shift in the probability distribution occurs and at the same time  we are interested in modelling the conditional distribution $p(\D|\sys)$ for a system not belonging to the class used in the training process.
    
    \item \textbf{Change of prediction task:} Adapt the meta-model trained  to perform a prediction task (for example, $n$-step ahead prediction) towards performing another prediction task (for example, $n'$-step ahead prediction). Indeed, when $n \ll n'$, training for $n$-step ahead is usually easier than training for $n'$-step ahead. Thus, training can be done first for with an $n$-step ahead objective. The harder $n'$-step ahead prediction objective can be employed at a later stage to refine the models through a few iterations of gradient descent or to update only a subset of the model's parameters. This is in line with the concept of \emph{curriculum learning}, where a learning problem is tackled by solving tasks of increasing difficulty~\citep{wang2021survey}.
\end{itemize}

Model learning and adaptation is discussed in the following section, and the importance of adaptation in the three scenarios mentioned above is shown in Section~\ref{sec:examples} through numerical examples.

\section{Meta system identification: algorithms} \label{sec:meta_sys}

\subsection{Architecture}
\begin{figure*}[!bt]
\centering
\includegraphics[width=.7\textwidth]{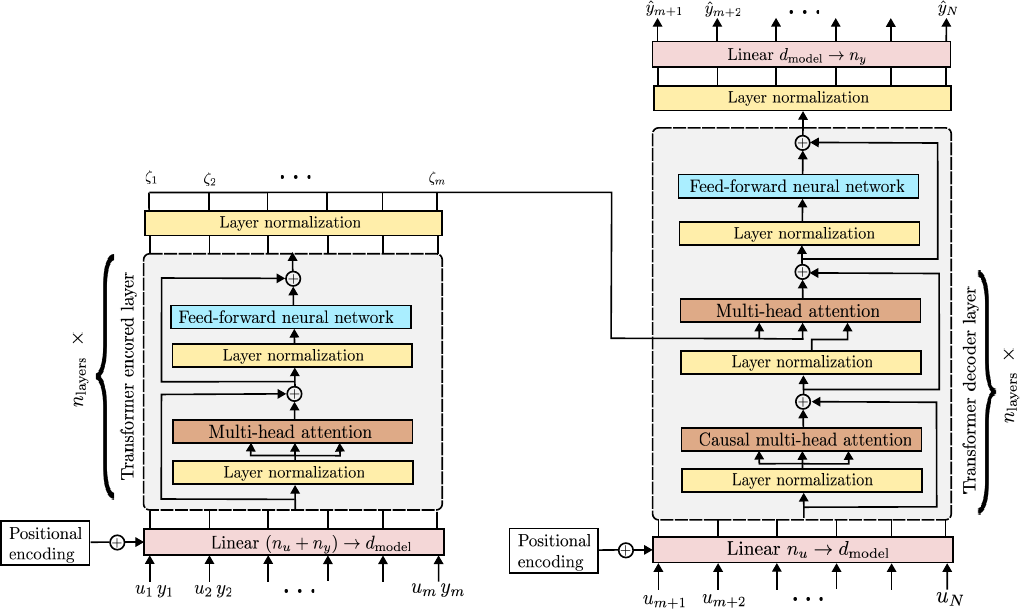}
\caption{Encoder-decoder Transformer  for multi-step-ahead simulation. 
With respect to the architecture in \citet{forgione2023context}, we use fixed positional encodings.}
\label{fig:encoder_decoder_arch}
\end{figure*}

The encoder-decoder Transformer architecture used as meta-model in this paper is illustrated in Fig.~\ref{fig:encoder_decoder_arch}. It is a rather standard Transformer architecture, similar to the one introduced in~\citet{vaswani2017attention} and adapted to receive as input/produce as output real-valued samples instead of categorical world tokens \citep{forgione2023context}. 

The   Transformer  is fully specified by the choice of the following hyper-parameters: number of layers $n_{\rm layers}$; number of units in each layer $d_{\rm model}$; number of heads $n_{\rm heads}$; and context window length $n_{\rm ctx}$. 

\subsection{Training} \label{sec:training}
 The encoder processes an input/output sequence  
$u_{1:m}, y_{1:m}$  and generates an embedding sequence $\zeta_{1:{m}}$, which is then processed by the decoder along with a test input  $u_{m+1:\nsamp}$ (the latter with causal restriction) to produce the 
sequence of predictions $\hat y_{m+1:\nsamp}$ up to step $\nsamp$. The   weights $\phi$ of the meta-model $\free_{\phi}$ (namely, the learning parameters of the Transformer)  are obtained by minimizing over $\phi$  the expected loss:
\begin{equation}
\label{eq:simulation_objective}
     J^* = \E_{p(\D)} 
    \left [ 
    \norm{y_{m+1:\nsamp} -  \free_\phi (u_{1:m}, y_{1:m}, u_{m+1:\nsamp})
    }^2
    \right ].
\end{equation}
The expected loss $J^*$ is approximated with an average $J$ over a finite number of sampled systems $\sys^{(i)}$ and  datasets $\D^{(i)}$ according to:
\begin{multline}
	\label{eq:simulation_objective_samples}
    J = 
    \frac{1}{b}
    \sum_{i=1}^b
    \norm{y_{m+1:\nsamp}^{(i)} - \free_\phi (u_{1:m}^{(i)}, y_{1:m}^{(i)}, u_{m+1:\nsamp}^{(i)})}^2
    ,
\end{multline}
where   $b$ denotes the number of randomly generated systems, each one providing a  input/output dataset  $\D^{(i)}$ of length $N$. 

The loss \eqref{eq:simulation_objective_samples} is then optimized through 
\emph{stochastic gradient descent},  with  $b$ new datasets resampled \emph{at each iteration}.

\subsection{Adaptation}

Once a nominal meta-model has been trained, 
an adaptation step is performed to refine its capabilities to solve new tasks. This is achieved by employing the pre-trained Transformer, with its learned weights $\phi$, as an initial condition for subsequent training phases on the new (possibly small-size) dataset, which is specific to the task we aim to describe. This approach leverages the knowledge already acquired by the model, allowing for a more efficient and targeted learning process. This additional training is conducted over a limited number of iterations, applying gradient updates to fine-tune the model's parameters specifically for the new data characteristics. To ensure the model's generalization ability and prevent overfitting,  an early stopping criterion is employed  based on the performance on a separate validation dataset. This method not only enhances the model's adaptability to new system dynamics, but also significantly reduces the computational resources and time required for retraining.

\section{Numerical examples} \label{sec:examples}

In this section, we present four numerical examples to demonstrate the effectiveness of meta-modelling and adaptation:
\begin{itemize}
    \item The first example does not consider adaptation, but illustrates  ``extrapolation'' properties of the meta-model across different classes. We show that the trained meta-models, even without fine-tuning, have a certain degree  of meta-generalization capabilities, namely, the ability to ``interpolate between classes'', enabling them to describe new classes not used in training.
    \item The second and third examples focus on meta-model adaptation. They address class-to-in-class-systems and class-to-out-of-class systems, respectively. The third example thus demonstrates the relevance of model adaptation in the case of a shift in the data-generating probability distribution $p(\mathcal{D})$. 
    \item The fourth example addresses refinement in the case of a task change, specifically transitioning from short-term to long-term predictions.
\end{itemize}

The \emph{AdamW} algorithm~\citep{loshchilov2017decoupled} is used to train the parameters $\phi$ of the  meta-model $\free_\phi$, by minimizing the empirical risk~\eqref{eq:simulation_objective_samples}.

The software has been developed in PyTorch and it is  available in the GitHub repository \url{https://github.com/forgi86/sysid-transformers-transfer}. Computations are performed on an Nvidia RTX 3090 GPU.

\subsection{Meta-generalization to unseen system classes}
\begin{figure}
\begin{center}
\includegraphics[width=.8\linewidth]{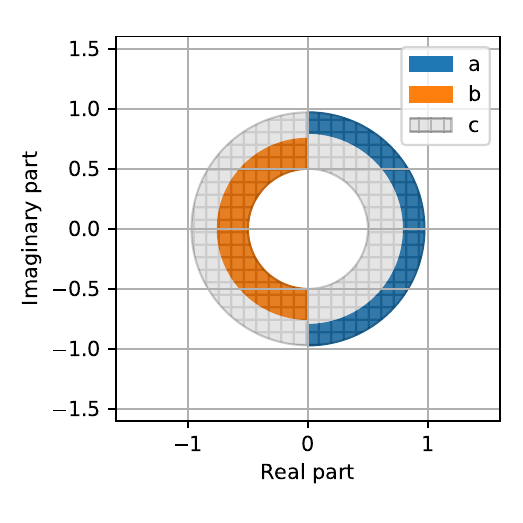}
\caption{Regions of the poles for the distributions $a$, $b$, and $c$ in the unit circle.}
\label{fig:pole_distribution}
\end{center}
\end{figure}

We illustrate the meta-generalization properties of the Transformer to a data distribution outside of the training region. To this end, we consider different data distributions generated by LTI systems with random order between 1 and 10, and poles constrained in different regions of the unit circle. In particular, let us denote for convenience $a, b, c$ the data distributions generated by linear systems  having poles uniformly distributed in the magnitude/phase regions of $(0.8, 0.97) / (-\pi/2, \pi/2)$, $(0.5, 0.75) / (\pi/2, 3/4\pi)$, and $(0.5, 0.97)/(-\pi, \pi)$, respectively.  Note that the pole region of $c$ is a superset of $a$ and $b$ (see Fig.~\ref{fig:pole_distribution}). 
Furthermore, let us denote by $a+b$ the mixture distribution obtained by sampling with equal probability from $a$ and $b$.

\begin{figure}[h]
    \begin{subfigure}{\linewidth}
    \includegraphics[width=\linewidth]{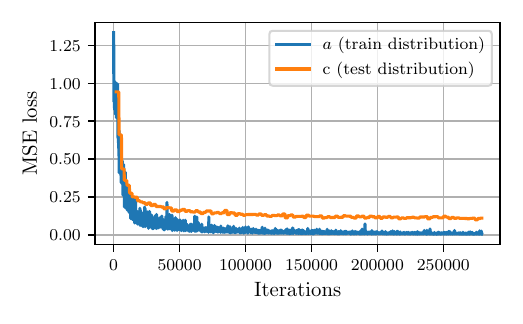}
    \end{subfigure}
    \begin{subfigure}{\linewidth}
    \includegraphics[width=\linewidth]{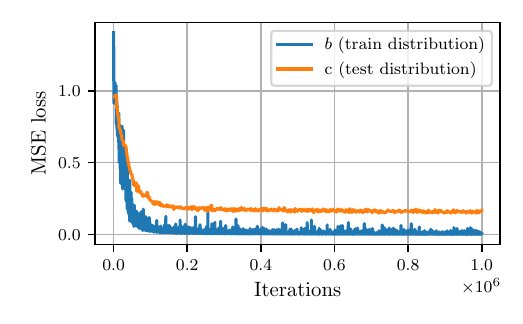}
    \end{subfigure}
    \begin{subfigure}{\linewidth}
    \includegraphics[width=\linewidth]{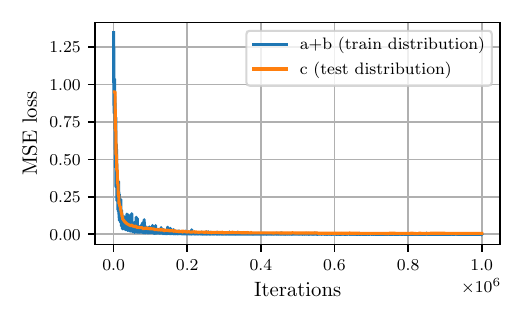}
    \end{subfigure}
\caption{Training and test loss. Training on $a$ (upper panel), on $b$ (middle panel), and on $(a, b)$ (bottom panel). Test is always performed on $c$.}
\label{fig:distribution_shift}
\end{figure}

We assess the generalization properties of a meta-model learned  by repeating the learning procedure on $a$, $b$, and $a+b$, while monitoring the performance obtained on $c$.
Results are visualized in Fig.~\ref{fig:distribution_shift}, where we report the training loss curves obtained by training on $a$, $b$, and $a+b$, together with the corresponding ``test'' loss, always evaluated on $c$. It is evident that, if we train on the data distributions $a$ or $b$, the meta generalization on $c$ is rather poor, i.e. there is a significant gap between the training performance on $a$ or $b$ and test performance measured on $c$ (top and middle panel). However, if we train on $a + c$, the meta generalization gap is much narrower (bottom right panel).
Intriguingly, when training on $a+b$, the Transformer appears  to be able to extrapolate effectively from magnitude/phase values that are only seen in one of the two distributions.

\subsection{Adaptation from class to system}
We consider a nominal meta-model $ \free_\phi$ trained according to the setting in \citet{forgione2023context} and describing the class of stable \emph{Wiener-Hammerstein} (WH) dynamical systems with structure $G_1$--$F$--$G_2$, where $G_1$, $G_2$ are SISO LTI blocks randomly generated with order comprised between 1 to 5, and $F$ is a static feed-forward neural network with one hidden layer and randomly generated  parameters.

The Transformer structure is defined by $n_{\rm layers}=12$
layers;  $d_{\rm model}=128$ units in each layer;  $n_{\rm heads}=4$ attention heads;  $m = 400$ and $n=N-m=100$ encoder's and decoder's context window length, respectively. The total number of weights characterizing the Transformer is $5.6$~M. Training was performed on $6$~M iterations, corresponding to a run time of 5.4 days. At each iteration, $b=32$ new  WH systems and 
 white Gaussian input sequences are  randomly sampled, and corresponding output sequences are simulated.

\subsubsection{In-class system}
\begin{figure}[t]
\begin{subfigure}{0.49\linewidth}
\includegraphics[width=\linewidth]{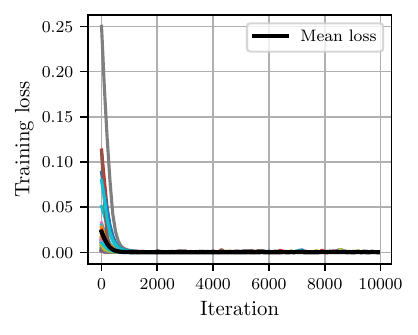}
\label{fig:WH_model_adaptation}   
\end{subfigure}
\hfill
\begin{subfigure}{0.49\linewidth}
\includegraphics[width=\linewidth]{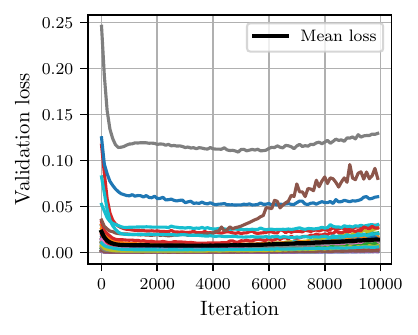}
\label{WH_model_adaptation_train_loss}
\end{subfigure}
\caption{WH model adaptation: training (left) and validation (right) over 50 Monte Carlo runs. The mean loss   
is reported as a thick black line.}
\label{fig:WH_model_adaptation_loss}
\end{figure}

\begin{figure}[t]
\centering
\includegraphics[width=.8\linewidth]{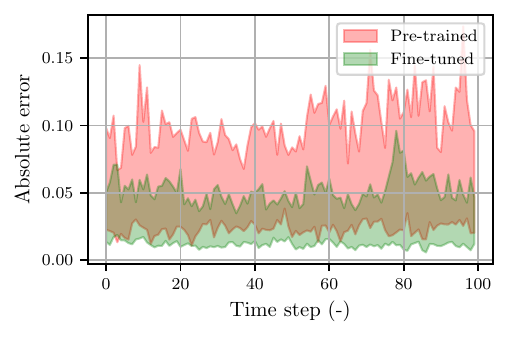}
\caption{WH model adaptation: area between the 25\textsuperscript{th} and the 75\textsuperscript{th} quantiles of the absolute error over time for the pre-trained (red) and fine-tuned (green) models. }
\label{fig:WH_model_adaptation_quantiles}   
\end{figure}

The nominal meta-model is then adapted to describe \emph{specific} WH systems. A Monte Carlo analysis is carried out over $50$ randomly sampled system. At each Monte Carlo run,  we sample one WH system and generate from that system $140$ input/output sequences, each one of length $N=500$. Out of these sequences, $100$ are used for training, $20$ for validation, and $20$ for test.  Model adaptation is performed on $10'000$ iterations, corresponding to about $20$ minutes per run.  Training and validation losses are reported in Fig.~\ref{fig:WH_model_adaptation_loss} \emph{vs.} number of iterations. We observe that, in average, the validation loss achieves its minimum in about $4'000$ iterations. Fig.~\ref{fig:WH_model_adaptation_quantiles}  shows the area between the 25\textsuperscript{th} and 75\textsuperscript{th} quantiles of the absolute value of the simulation error over test data and for each simulation time step. For comparison, the  quantiles corresponding to the pre-trained (non fine-tuned) meta-model are reported in the same figure, showing the effectiveness of the adaptation procedure.

\subsubsection{Out-of-class system}
In this numerical example, we adapt the nominal meta-model $\free_\phi$ (pre-trained on the WH class) to describe \emph{Parallel Wiener-Hammerstein} (PWH) systems. The same experimental settings considered in the previous example are adopted. 

\begin{figure}[t]
\begin{subfigure}{0.49\linewidth}
\includegraphics[width=\linewidth]{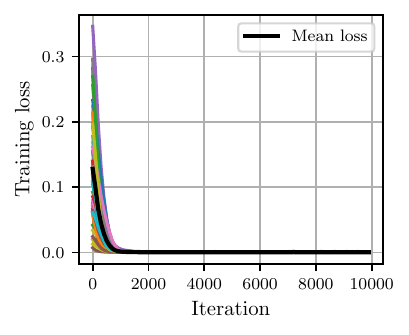}
\label{fig:PWH_model_adaptation_train_loss}   
\end{subfigure}
\hfill
\begin{subfigure}{0.49\linewidth}
\includegraphics[width=\linewidth]{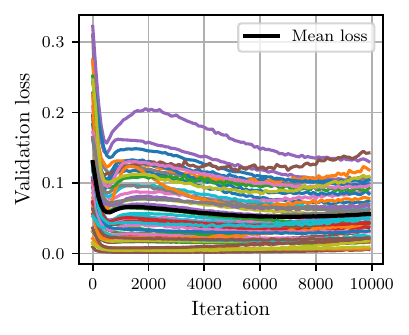}
\label{fig:PWH_model_adaptation_val_loss}
\end{subfigure}
\caption{PWH model adaptation: training (left) and validation (right) losses over 50 Monte Carlo runs. The mean loss  
is reported as a thick black line.}
\label{fig:PWH_model_adaptation_loss}
\end{figure}

\begin{figure}[t]
\centering
\includegraphics[width=.8\linewidth]{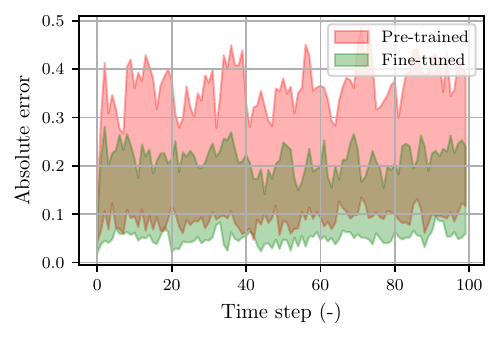}
\caption{PWH model adaptation: area between the 25th and the 75th quantiles of the absolute error over time for the pre-trained (red) and fine-tuned (green) models. }
\label{fig:PWH_model_adaptation_quantiles}   
\end{figure}

Training and validation losses are reported in Fig.~\ref{fig:PWH_model_adaptation_loss} \emph{vs.} the iteration index. We observe that, in average, the validation loss achieves its minimum in about $6'000$ iterations. Fig.~\ref{fig:PWH_model_adaptation_quantiles}  shows the area between the 25\textsuperscript{th} and 75\textsuperscript{th} quantiles of the absolute value of the simulation error over test data and for each simulation time step. For comparison, the  quantiles corresponding to the pre-trained (non fine-tuned) meta-model are reported in the same figure. The reported results show the capabilities of the meta-model to cope  (after fine-tuning) with data distribution shifts. 

\subsection{Short-to-long simulation}
\begin{figure}
\centering
\includegraphics{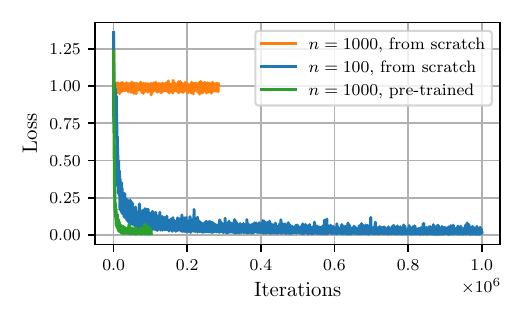}
\caption{Training loss of the Transformer for short-term simulation ($n=100$) initialized from scratch (blue); for long-term simulation ($n=1000$) initialized from scratch (orange) and for long-term simulation ($n=1000$) with initialization from the weights trained on short-term sequences (green). In all cases, the context length is $m=400$.}
\label{fig:training_short_long}
\end{figure}

We consider the problem of meta-modelling the WH class to make predictions over the longer horizon of $n=1000$ steps.
We set-up training of the Transformer with the same configurations used in the previous cases. 
Fig.~\ref{fig:training_short_long} (orange curve) shows the training loss vs. the optimization iteration.
The loss is stuck around the numerical value 1.0, which corresponds to no meaningful learning.\footnote{Signals are normalized to zero mean and unit variance. A unit loss is thus achieved by a (dummy) predictor returning the zero constant.} The training procedure was thus interrupted after about 250'000 iterations where no progress occurred. 

Conversely, training from scratch a $n$-step-ahead simulator with $n=100$ was found to be feasible (blue curve), with 
the loss decreasing to a very low value after 1'000'000 iterations (approximately 1 day), similarly to the results reported in \citet{forgione2023context}. The trained meta-model obtains an average $\rm rmse$ index of 0.103. We then executed the optimization over the longer horizon $n=1000$ using the weights previously learned on the $n=100$ case as initialization, instead of starting from random ones. Interestingly, with this initialization (green curve), the loss decreased rapidly in the first 20'000 iterations and reached a plateau after about 80'000 iterations (approximately 21 hours), when the experiment was interrupted. The adapted meta-model achieves an average simulation $\rm rmse$ of $0.113$ over $1000$ steps.

\section{Conclusion} \label{sec:conc}
The paper demonstrates the potential of in-context learning for system identification, emphasizing the importance of model adaptability. It showcases how meta-modelling can extrapolate between different classes, enabling the description of new, unseen system classes. The key scenarios explored include refining meta-models for specific systems, adapting to behaviours outside the initial training class, and transitioning models for new prediction tasks. The adaptation process is shown to require limited computational time and data compared to meta-model pretraining, with an early stopping criterion ensuring model generalization and preventing overfitting.

This work aims to open research directions for new learning and modelling paradigms in system identification, leveraging the power of neural network architectures like Transformers and by exploiting the shared features among similar dynamical systems to identify new systems of interest with a limited amount of data and time. 

\begin{ack}
The activities of Marco Forgione have  been supported by HASLER STIFTUNG under the project QUACK: QUAntifying the unCertainty of dynamical neural networKs.
\end{ack}

\bibliography{biblio}             

\end{document}